\documentclass[sigconf]{acmart}

\usepackage{xcolor}
\usepackage{ifthen}
\newcommand{\markupdraft}[2]{
    \ifthenelse{\equal{#1}{display}}{#2}{}
    \ifthenelse{\equal{#1}{color}}{\color{#2}}{}
}

\newcommand{\newcolored}[3][]{{\markupdraft{color}{#2}#3}
    \ifthenelse{\equal{#1}{}}{}{\markupdraft{display}{{\color{yellow!70!black}[#1]}}}}
\newcommand{\del}[2][]{{\markupdraft{display}{{\color{orange}[removed: ``#2''[#1]]}}}} 
\newcommand{\new}[2][]{\newcolored[#1]{blue}{#2}}
\newcommand{\nnew}[2][]{\newcolored[#1]{red}{#2}}

\renewcommand{\del}[2]{}  
\renewcommand{\markupdraft}[2]{}  

\newcommand{\E}{\mathbb{E}}
\newcommand{\R}{\mathbb{R}}
\newcommand{\X}{\mathcal{X}}

\newcommand{\I}[1][\dime]{\,\mathrm{I}_{#1}}
\newcommand{\T}{\mathrm{T}}
\newcommand{\N}{\mathcal{N}}

\newcommand{\x}{\boldsymbol{x}}
\newcommand{\y}{\boldsymbol{y}}
\newcommand{\z}{\boldsymbol{\xi}}
\newcommand{\vb}{\boldsymbol{v}}

\newcommand{\mv}{\boldsymbol{m}}
\newcommand{\cov}{\boldsymbol{C}}

\newcommand{\p}{\boldsymbol{p}}
\newcommand{\A}{\boldsymbol{A}}

\newcommand{\ele}[1]{[ #1 ]}
\newcommand{\dele}[1]{\langle #1 \rangle}

\newcommand{\mueff}{\mu_{\mathrm{w}}}
\newcommand{\enc}{\textsc{Enc}}

\newcommand{\ci}{\mathrm{CI}}
\newcommand{\plow}{p_\mathrm{low}}
\newcommand{\pup}{p_\mathrm{up}}
\newcommand{\pmid}{p_\mathrm{mid}}

\newcommand{\psucc}{p_\mathrm{succ}}
\newcommand{\ptarget}{p_\mathrm{target}}

\usepackage[ruled]{algorithm2e} 
\usepackage{algorithmic}

\AtBeginDocument{%
  \providecommand\BibTeX{{%
    \normalfont B\kern-0.5em{\scshape i\kern-0.25em b}\kern-0.8em\TeX}}}

\copyrightyear{2023} 
\acmYear{2023} 
\setcopyright{acmlicensed}\acmConference[GECCO '23]{Genetic and Evolutionary Computation Conference}{July 15--19, 2023}{Lisbon, Portugal}
\acmBooktitle{Genetic and Evolutionary Computation Conference (GECCO '23), July 15--19, 2023, Lisbon, Portugal}
\acmPrice{15.00}
\acmDOI{10.1145/3583131.3590516}
\acmISBN{979-8-4007-0119-1/23/07}

\begin{document}

\title[(1+1)-CMA-ES with Margin for Discrete and Mixed-Integer Problems]{(1+1)-CMA-ES with Margin\\for Discrete and Mixed-Integer Problems}

\author{Yohei Watanabe}
\email{watanabe-yohei-jc@ynu.jp}
\orcid{0000-0002-9258-5258}
\affiliation{%
  \institution{Yokohama National University}
  \city{Yokohama}
  \state{Kanagawa}
  \country{Japan}
  \postcode{240-8501}
}

\author{Kento Uchida}
\email{kento.u.hnr1401@gmail.com}
\orcid{0000-0002-4179-6020}
\affiliation{%
  \institution{Yokohama National University}
  \city{Yokohama}
  \state{Kanagawa}
  \country{Japan}
  \postcode{240-8501}
}

\author{Ryoki Hamano}
\email{hamano-ryoki-pd@ynu.jp}
\orcid{0000-0002-4425-1683}
\affiliation{%
  \institution{Yokohama National University}
  \city{Yokohama}
  \state{Kanagawa}
  \country{Japan}
  \postcode{240-8501}
}

\author{Shota Saito}
\email{saito-shota-bt@ynu.jp}
\orcid{0000-0002-9863-6765}
\affiliation{%
  \institution{Yokohama National University \and SkillUp AI Co., Ltd.}
  \city{Yokohama}
  \state{Kanagawa}
  \country{Japan}
  \postcode{240-8501}
}

\author{Masahiro Nomura}
\email{nomura\_masahiro@cyberagent.co.jp}
\orcid{0000-0002-4945-5984}
\affiliation{%
  \institution{CyberAgent, Inc.}
  \city{Shibuya}
  \state{Tokyo}
  \country{Japan}
  \postcode{150-0042}
}

\author{Shinichi Shirakawa}
\email{shirakawa-shinichi-bg@ynu.ac.jp}
\orcid{0000-0002-4659-6108}
\affiliation{%
  \institution{Yokohama National University}
  \city{Yokohama}
  \state{Kanagawa}
  \country{Japan}
  \postcode{240-8501}
}

\renewcommand{\shortauthors}{Y. Watanabe et al.}

\begin{abstract}
The covariance matrix adaptation evolution strategy (CMA-ES) is an efficient continuous black-box optimization method.
The CMA-ES possesses many attractive features, including invariance properties and a well-tuned default hyperparameter setting. Moreover, several components to specialize the CMA-ES have been proposed, such as noise handling and constraint handling. To utilize these advantages in mixed-integer optimization problems, the CMA-ES with margin has been proposed. The CMA-ES with margin prevents the premature convergence of discrete variables by the margin correction, in which the distribution parameters are modified to leave the generation probability for changing the discrete variable. 
The margin correction has been applied to ($\mu/\mu_\mathrm{w}$,$\lambda$)-CMA-ES, while this paper introduces the margin correction into (1+1)-CMA-ES, an elitist version of CMA-ES. \del{The (1+1)-CMA-ES is advantageous in unimodal functions and requires less computational cost in high-dimensional problems.}{}\new{The (1+1)-CMA-ES is often advantageous for unimodal functions and can be computationally less expensive.}
To tackle the performance deterioration on mixed-integer optimization, we use the discretized elitist solution as the mean of the sampling distribution and modify the margin correction not to move the elitist solution. 
The numerical simulation using benchmark functions on mixed-integer, integer, and binary domains shows that (1+1)-CMA-ES with margin outperforms the CMA-ES with margin and is better than or comparable with several specialized methods to a particular search domain. 
\end{abstract}




\maketitle

\section{Introduction}

\paragraph{Backgrounds}
The covariance matrix adaptation evolution strategy (CMA-ES)~\cite{hansen:1996:cmaes,hansen:2003:ec} is an efficient optimization method in continuous black-box optimizations. The CMA-ES generates candidate solutions using a multivariate Gaussian distribution and performs black-box optimization by iteratively updating the distribution parameters. The CMA-ES possesses several attractive features. The well-tuned default hyperparameter setting~\cite{hansen:2011:tutorial} makes the CMA-ES quasi-hyperparameter-free, which does not require the cost of hyperparameter tuning. Owing to the invariance properties, the CMA-ES works well on a wide range of problems, such as non-separable and/or ill-conditioned problems. As another advantage, several components specialized for the CMA-ES have been proposed, such as noise handling~\cite{cmaes:noise1,cmaes:noise2}, constraint handling~\cite{cmaes:noise1,cmaes:constraint2,cmaes:constraint3}, and multi-objective CMA-ES~\cite{cmaes:multiobj1}.

In the field of black-box optimization, there are several kinds of domains of the design variables, such as continuous, integer, and binary domains. The mixed-integer optimization problems contain both continuous and discrete variables.
Many studies have presented advanced black-box optimization methods\new{~\cite{Rios:2013, larson_menickelly_wild_2019}}. However, most of them focus on only a part of the kinds of domains, especially the continuous and binary domains. In particular, the integer and mixed-integer optimization methods have not been investigated actively, even though there are a lot of real-world applications in these domains~\cite{integer-opt:example1,mixed-integer-opt:example1,mixed-integer-opt:example2,mixed-integer-opt:example3}.

\emph{The CMA-ES with margin}~\cite{cmaeswm} is an efficient mixed-integer optimization method that can inherit the advantages of the CMA-ES. To prevent the premature convergence of discrete variables, the CMA-ES with margin introduces a lower bound on the marginal probability, referred to as \emph{the margin}, so that the samples are not fixed to a single discrete value. The CMA-ES with margin applies an affine transformation, called \emph{the margin correction}, to ensure the margin.
The excellent performance of the CMA-ES with margin on the mixed-integer domain is confirmed in~\cite{cmaeswm}. In principle, the CMA-ES with margin can be applied to the integer and binary optimization problems by setting the number of continuous variables to zero, where the performance in those cases has yet to be investigated.

\paragraph{Contributions}
This paper introduces the margin correction of the CMA-ES with margin into the (1+1)-CMA-ES~\cite{Igel:2006,Suttorp:2009}, an elitist version of CMA-ES, and proposes {\it the (1+1)-CMA-ES with margin}. The (1+1)-CMA-ES shows powerful optimization performance on unimodal functions in the continuous domain. 
\del{When applying the original margin correction to the (1+1)-CMA-ES, it fails to optimize the mixed-integer problem due to the premature convergence of the continuous variables. To tackle this issue, we introduce the discretization of \del{the mean vector of the Gaussian distribution}{}\new{the elements of the mean vector corresponding to discrete variables} and modify the margin correction not to move the mean vector.}{}
We introduce a revised update rule of the mean vector that prevents the premature convergence on mixed-integer problems and modify the margin correction not to move the mean vector.
We also propose a post-process that modifies the updated distribution parameters so that the behavior is not affected by the numerical errors in binary and integer optimizations.

The numerical simulation using benchmark functions on mixed-integer, integer, and binary domains shows that the (1+1)-CMA-ES with margin outperforms the CMA-ES with margin. Moreover, compared to optimization methods designed solely for binary optimization, we found that the (1+1)-CMA-ES with margin outperforms the compact genetic algorithm~\cite{cga} and population-based incremental learning~\cite{pbil}, and is comparable with (1+1)-EA. The experimental results show the potential of CMA-ES with margin as a universal optimizer for various variable-type problems.
This study provides new possibilities for developing discrete and mixed-integer optimization methods derived from the CMA-ES.

\paragraph{Notations}
We denote the $j$-th element of a vector $\boldsymbol{a}$ and $j$-th diagonal element of a matrix $\boldsymbol{A}$ as $\ele{\boldsymbol{a}}_j$ and $\dele{\boldsymbol{A}}_j$, respectively. The identity matrix is denoted as $\I \in \R^{N \times N}$.

\section{CMA-ES with Margin} \label{sec:cmaeswm}
We consider the mixed-integer minimization problem of an objective function $f$ whose first $N_\mathrm{co}$ design variables are continuous variables, and the rest $N_\mathrm{in}$ variables are integer (or binary) variables. The total number of dimensions is $N = N_\mathrm{co}+N_\mathrm{in}$. 
For $j = N_\mathrm{co} + 1, \cdots, N$, the set of possible values for the $j$-th design variable is given by $\mathcal{Z}_j = \{ z_{j,1}, \cdots z_{j, K_j} \}$. 
We assume $z_{j,k}$ to be the $k$-th smallest value in $\mathcal{Z}_j$ without loss of generality. 
Totally, the search space is given by $\X := \R^{N_\mathrm{co}} \times \mathcal{Z}_{N_\mathrm{co} + 1} \times \cdots \times \mathcal{Z}_N$.

The CMA-ES with margin~\cite{cmaeswm} employs a multivariate Gaussian distribution $\N(\mv^{(t)}, (\sigma^{(t)})^2 \cov^{(t)})$ parameterized by the mean vector $\mv^{(t)} \in \R^N$, covariance matrix $\cov^{(t)} \in \R^{N \times N}$, and step-size $\sigma^{(t)} \in \R_{>0}$.
The CMA-ES with margin also contains a diagonal matrix $\A^{(t)} \in \R^{N \times N}$, that is initialized as $\A^{(0)} = \I$.
The single update of the CMA-ES with margin consists of two components; the same update procedure of the distribution parameters as the original CMA-ES and the margin correction. The pseudocode of the CMA-ES with margin is shown in Algorithm~\ref{alg:cmaeswm}.

\subsection{Update of Distribution Parameters} \label{sec:cmaeswm-update}
In each iteration $t$, the CMA-ES with margin generates $\lambda$ candidate solutions $\x_1, \cdots, \x_\lambda \in \R^N$ and affine transformed solutions $\vb_1, \cdots, \vb_\lambda \in \R^N$ as
\begin{align}
    \y_i &= ( \cov^{(t)} )^{\frac{1}{2}} \z_i \\
    \x_i &= \mv^{(t)} + \sigma^{(t)} \y_i \label{eq:cmaeswm-sample-x} \\
    \vb_i &= \mv^{(t)} + \sigma^{(t)} \A^{(t)} \y_i \enspace,
\end{align}
where $\z_1, \cdots, \z_\lambda$ are independent and identically distributed (i.i.d.) samples generated from the $N$-dimensional standard Gaussian distribution $\N(\mathbf{0}, \I)$ and $( \cov^{(t)} )^{\frac{1}{2}}$ is the square root of the covariance matrix $\cov^{(t)}$. 

Then, the affine transformed solutions are transformed into $\bar{\vb}_1, \cdots, \bar{\vb}_\lambda \in \X$ by the encoding function $\enc:\R^N \to \mathcal{X}$ to be evaluated on the objective function $f$.
The elements corresponding to continuous variables are unchanged, i.e., $\ele{\bar{\vb}_i}_j = \ele{\enc(\vb_i)}_j = \ele{\vb_i}_j$ for $j = 1, \cdots, N_\mathrm{co}$. 
For $j = N_\mathrm{co}+1, \cdots, N$, the $j$-th element of $\bar{\vb}_i$ is given by
\begin{align}
    \ele{\bar{\vb}_i}_j = \ele{\enc(\vb_i)}_j = \begin{cases}
        z_{j,1} & \text{if} \quad \ele{\vb_i}_j \leq \ell_{j, 1|2} \\
        z_{j,k} & \text{if} \quad \ell_{j, k-1|k} < \ele{\vb_i}_j \leq \ell_{j, k|k+1} \\
        z_{j,K_j} & \text{if} \quad \ell_{j, K_j-1|K_j} < \ele{\vb_i}_j 
    \end{cases} \enspace,
    \label{eq:enc}
\end{align}
where $\ell_{j, k|k+1}$ is the midpoint of $z_{j, k}$ and $z_{j, k+1}$, i.e., $\ell_{j, k|k+1} = (z_{j, k} + z_{j, k+1}) / 2$.
We denote the index of $i$-th best sample as $i:\lambda$, which satisfies $f(\bar{\vb}_{1:\lambda}) \leq \cdots \leq f(\bar{\vb}_{\lambda:\lambda})$.

\begin{algorithm}[t] 
\caption{The CMA-ES with margin}
\begin{algorithmic}[1] \label{alg:cmaeswm}
\REQUIRE The objective function $f$ to be optimized
\REQUIRE $\mv^{(0)}, \cov^{(0)}, \sigma^{(0)}, \A^{(0)}$ 
\WHILE{termination conditions are not met}
\FOR{$i = 1$ to $\lambda$}
\STATE Generate $\y_i = ( \cov^{(t)} )^{\frac{1}{2}} \z_i$ with $\z_i \sim \N(\mathbf{0}, \I)$. 
\STATE Compute $\x_i = \mv^{(t)} + \sigma^{(t)} \y_i$.
\STATE Compute $\vb_i = \mv^{(t)} + \sigma^{(t)} \A^{(t)} \y_i$.
\STATE Discretize $\vb_i$ as $\bar{\vb}_i = \enc(\vb_i)$.
\STATE Evaluate $f( \bar{\vb}_i )$.
\ENDFOR
\STATE Update $\mv^{(t)}, \p_\sigma^{(t)}, \p_c^{(t)}, \cov^{(t)}$ and $\sigma^{(t)}$.
\STATE Modify $\mv^{(t+1)}$ and $\A^{(t+1)}$ by margin correction.
\STATE $t \leftarrow t+1$
\ENDWHILE
\end{algorithmic} 
\end{algorithm}

Then, the distribution parameters are updated based on the ranking of candidate solutions. Introducing the weights $w_1, \cdots, w_\lambda$ satisfying \new{$w_1 \geq \cdots \geq w_\mu > 0 \geq w_{\mu+1} \geq \cdots \geq w_\lambda$} and $\sum^\mu_{i=1} w_i = 1$ \new{for $\mu \leq \lambda$}, the mean vector is updated as
\begin{align}
    \mv^{(t+1)} = \mv^{(t)} + c_m \sum_{i=1}^{\mu} w_i ( \x_{i:\lambda} - \mv^{(t)} ) \enspace, 
    \label{eq:mean-update} 
\end{align}
where $c_m > 0$ is the learning rate.
In the update rule of the covariance matrix and step-size, two evolution paths $\p_c^{(t)} \in \R^N$ and $\p_\sigma^{(t)} \in \R^N$ are used. They are initialized as $\p_c^{(0)} = \p_\sigma^{(0)} = \mathbf{0}$ and updated as
\begin{align}
    \p_\sigma^{(t+1)} &= (1 - c_\sigma) \p_\sigma^{(t)} + \sqrt{c_\sigma (2-c_\sigma) \mueff} \sum_{i=1}^{\mu} w_i \z_{i:\lambda} 
    \label{eq:csa-path-update} \\
    \p_c^{(t+1)} &= (1 - c_c) \p_c^{(t)} + h_\sigma^{(t+1)} \sqrt{c_c (2-c_c) \mueff} \sum_{i=1}^{\mu} w_i \y_{i:\lambda} \enspace, 
    \label{eq:rank-one-path-update}
\end{align}
where $c_\sigma > 0$ and $c_c > 0$ are the cumulative rates, and $\mueff = ( \sum_{i=1}^{\mu} w_i^2 )^{-1}$ is the variance effective selection mass. The Heaviside function takes $h_\sigma^{(t+1)} = 1$ if it holds
\begin{align}
    \frac{ \|\p_\sigma^{(t+1)}\| }{\sqrt{ 1 - (1 - c_\sigma )^{2 (t + 1)} }} < \left( 1.4 + \frac{2}{N + 1} \right) \E \left[ \| \mathcal{N}(\mathbf{0}, \I) \| \right] \enspace, 
    \label{eq:csa-heaviside}
\end{align}
and it takes $h_\sigma^{(t+1)} = 0$ otherwise. The Heaviside function stalls the update of the evolution path $\p_c^{(t)}$ when the step-size increases dramatically. Then the covariance matrix is updated as
\begin{multline}
    \cov^{(t+1)} = \left(1 - c_\mu \sum_{i=1}^{\lambda} w_i - c_1 + (1 - h_\sigma^{(t+1)}) c_1 c_c (2 - c_c) \right) \cov^{(t)} \\
    + c_\mu \sum_{i=1}^{\lambda} w_i^\circ \y_{i:\lambda} \y_{i:\lambda}^\T + c_1 \p_c^{(t+1)} (\p_c^{(t+1)})^\T \enspace,
    \label{eq:cov-update}
\end{multline}
where $w_i^\circ$ is given by $w_i^\circ = w_i$ if $w_i \geq 0$, and $w_i^\circ = w_i \cdot N / \| \z_{i:\lambda} \|^2$ otherwise. The update rule of the step-size is
\begin{align}
    \sigma^{(t+1)} = \sigma^{(t)} \exp\left(\frac{c_\sigma}{d_\sigma}\left(\frac{\|\p_\sigma^{(t+1)}\|}{\E[\|\N(\mathbf{0}, \I)\|]}-1\right)\right) \enspace, 
    \label{eq:csa}
\end{align}
where $d_\sigma > 0$ is the damping factor.

\subsection{Margin Correction} \label{sec:cmaeswm-correction}
After the update of the distribution parameters $\mv^{(t)}$, $\cov^{(t)}$ and $\sigma^{(t)}$, the CMA-ES with margin modifies the updated mean vector $\mv^{(t+1)}$ and the diagonal elements of $\A^{(t)}$ corresponding to the integer (or binary) variables. 
This modification maintains the probability of not generating the integer $\ele{\bar{\mv}^{(t+1)}}_j := \ele{\enc( \mv^{(t+1)} )}_j$ nearest to the $j$-th element of mean vector larger than the margin parameter $\alpha$, i.e., $\Pr( \ele{\bar{\vb}}_j \neq \ele{\bar{\mv}^{(t+1)}}_j ) \geq \alpha$. The modification process is determined by whether $\ele{\bar{\mv}^{(t+1)}}_j$ equals to the smallest integer $z_{j,1}$ or largest integer $z_{j,K_j}$, or others.

\paragraph{Case of Smallest or Largest Integer}
When $\ele{\bar{\mv}^{(t+1)}}_j$ equals to $z_{j,1}$ or $z_{j,K_j}$, the CMA-ES with margin modifies the mean vector as follows. 
Let us define $\ci^{(t+1)}_{j, \gamma}$ with $\gamma \in (0,1)$ as 
\begin{align}
\ci^{(t+1)}_{j, \gamma} := \sqrt{ \chi^2_\mathrm{ppf}(\gamma) (\sigma^{(t+1)})^2 \dele{ \A^{(t)} \cov^{(t+1)} \A^{(t)} }_j } \enspace,
\end{align}
where $\chi^2_\mathrm{ppf}(\gamma)$ is $\gamma$-quantile of $\chi^2$-distribution with $1$ degree of freedom.
Then, the confidence interval of the probability $1-2\alpha$ for the $j$-th element of sample $\ele{\vb}_j$ becomes
\begin{align}
\left[ \ele{\mv^{(t+1)}}_j - \ci^{(t+1)}_{j, 1-2\alpha}, \ele{\mv^{(t+1)}}_j + \ci^{(t+1)}_{j, 1-2\alpha} \right] \enspace. \label{eq:cmaeswm:conf-interval}
\end{align}
\del{ When the nearest midpoint $\ell( \ele{\mv^{(t+1)}}_j ) \in \{ \ell_{1|2}, \ell_{K_j-1|K_j} \}$ to $j$-th element of the mean vector is not in this confidence interval, the $j$-th element is modified to the edge of the interval.}{} 
The CMA-ES with margin modifies the elements of the mean vector so that the midpoints exist in the confidence intervals.
Consequently, the modification of the $j$-th element of the mean vector reads
\begin{multline}
\ele{ \mv^{(t+1)} }_j \leftarrow \ell\left(\ele{ \mv^{(t+1)} }_j \right) + \mathrm{sign}\left( \ele{ \mv^{(t+1)} }_j - \ell\left( \ele{ \mv^{(t+1)} }_j \right) \right) \\
\cdot \min\left\{ \left| \ele{ \mv^{(t+1)} }_j - \ell\left( \ele{ \mv^{(t+1)} }_j \right) \right|, \ci^{(t+1)}_{j, 1-2\alpha} \right\} \enspace,
\end{multline}
where $\ell( \ele{\mv^{(t+1)}}_j ) \in \{ \ell_{1|2}, \ell_{K_j-1|K_j} \}$ is the nearest midpoint to $j$-th element of the mean vector before the modification.
In this case, the $j$-th diagonal element of $\A^{(t)}$ is not changed, i.e., $\dele{\A^{(t+1)}}_j = \dele{\A^{(t)}}_j$.
We note that the elements of the mean vector corresponding to the binary variables are modified by the above-mentioned modification process.

\paragraph{Case of Other Integers}
When $\ele{\bar{\mv}^{(t+1)}}_j$ is in $\{z_{j,2}, \cdots, z_{j,K_j-1}\}$, the CMA-ES with margin modifies $\ele{\mv^{(t+1)}}_j$ and $\dele{\A^{(t)}}_j$ as follows. Let us denote the nearest two midpoints to $\ele{\mv^{(t+1)}}_j$ as
\begin{align}
\ell_{\mathrm{low},j}^{(t+1)} &:= \max\left\{ l \in \left\{ \ell_{j,k|k+1} \right\}_{k=1, \cdots, K_j-1} : l < \ele{\mv^{(t+1)}}_j \right\} \enspace \\
\ell_{\mathrm{up},j}^{(t+1)} &:= \min\left\{ l \in \left\{ \ell_{j,k|k+1} \right\}_{k=1, \cdots, K_j-1} : \ele{\mv^{(t+1)}}_j \leq l \right\} \enspace.
\end{align}
The modification aims that both of 
\begin{align}
\plow &:= \Pr\left( \ele{\vb}_j \leq \ell_{\mathrm{low},j}^{(t+1)} \right) \qquad \text{and} \label{eq:cmeswm-margin:plow} \\
\pup &:= \Pr\left( \ell_{\mathrm{up},j}^{(t+1)} < \ele{\vb}_j \right) \label{eq:cmeswm-margin:pup}
\end{align}
are maintained above $\alpha / 2$ after the margin correction. 

As the first step of the margin correction, with $\pmid := 1 - \plow - \pup$, the corrected marginal probabilities are calculated as
\begin{align}
\plow' &= \max\{ \alpha / 2, \plow \} \\
\pup' &= \max\{ \alpha / 2, \pup \} \\
\plow'' &=  \plow' + \frac{ 1 - \plow' - \pup' - \pmid }{ \plow' + \pup' + \pmid - 3 \cdot \alpha / 2 } \left( \plow' - \frac{\alpha}{2} \right) \\
\pup'' &=  \pup' + \frac{ 1 - \plow' - \pup' - \pmid }{ \plow' + \pup' + \pmid - 3 \cdot \alpha / 2 } \left( \pup' - \frac{\alpha}{2} \right) \enspace.
\end{align}
This maintains $\plow'' \geq \alpha / 2$ and $\pup'' \geq \alpha / 2$. Then $\ele{\mv^{(t+1)}}_j$ and $\dele{\A^{(t)}}_j$ are modified to satisfy
\begin{align}
\Pr\left( \ele{\vb}_j \leq \ell_{\mathrm{low},j}^{(t+1)} \right) &= \plow'' \geq \frac{\alpha}{2} \qquad \text{and} \\
\Pr\left( \ell_{\mathrm{up},j}^{(t+1)} < \ele{\vb}_j \right) &= \pup'' \geq \frac{\alpha}{2}
\end{align}
after the margin correction. To achieve this, $\ele{\mv^{(t+1)}}_j$ and $\dele{\A^{(t)}}_j$ are modified as
\begin{align}
\ele{\mv^{(t+1)}}_j &\leftarrow \frac{ \ell_{\mathrm{low},j}^{(t+1)} \sqrt{q_\mathrm{up}} + \ell_{\mathrm{up},j}^{(t+1)} \sqrt{q_\mathrm{low}} }{\sqrt{q_\mathrm{up}} + \sqrt{q_\mathrm{low}}} \label{eq:cmeswm-margin:mean} \\
\dele{\A^{(t+1)}}_j &\leftarrow \frac{ \ell_{\mathrm{up},j}^{(t+1)} - \ell_{\mathrm{low},j}^{(t+1)} }{ \sigma^{(t+1)} \sqrt{ \dele{ \cov^{(t+1)} }_j } \left( \sqrt{q_\mathrm{up}} + \sqrt{q_\mathrm{low}} \right) } \enspace, \label{eq:cmeswm-margin:A}
\end{align}
where $q_\mathrm{low} = \chi^2_\mathrm{ppf}(1 - 2 \plow'')$ and $q_\mathrm{up} = \chi^2_\mathrm{ppf}(1 - 2 \pup'')$. 

\subsection{Default Hyperparameter Setting} \label{sec:cmaeswm-hyperparameter}
The CMA-ES with margin shares the default hyperparameter setting with the CMA-ES proposed in~\cite{hansen:2011:tutorial} except for the margin parameter $\alpha$. The default setting of the margin parameter $\alpha = (\lambda N)^{-1}$ is tuned by numerical simulations. 
This default hyperparameter setting makes the CMA-ES with margin a quasi-hyperparameter-free optimization method that does not require hyperparameter tuning.


\section{(1+1)-CMA-ES with Margin} \label{sec:ecmaeswm}
We propose a variant of CMA-ES with margin combined with the elitist strategy, termed {\it (1+1)-CMA-ES with margin}. The proposed method is obtained by introducing the margin correction in Section~\ref{sec:cmaeswm-correction} to the (1+1)-CMA-ES~\cite{Suttorp:2009}. Algorithm~\ref{alg:ecmaeswm} shows the optimization process of the (1+1)-CMA-ES with margin.

\subsection{Overall Procedure} \label{sec:ecmaeswm-overall}
First, the (1+1)-CMA-ES with margin generates a sample as
\begin{align}
    \y_\mathrm{new} &= ( \cov^{(t)} )^{\frac{1}{2}} \z \\
    \vb_\mathrm{new} &= \mv^{(t)} + \sigma^{(t)} \A^{(t)} \y_\mathrm{new} \enspace,
\end{align}
where $\z \sim \N(\mathbf{0}, \I)$. 
Unlike in the CMA-ES with margin, the sample \new{in~\eqref{eq:cmaeswm-sample-x},} whose law is given by $\N(\mv^{(t)}, (\sigma^{(t)})^2 \cov^{(t)})$\new{,} is not computed because the (1+1)-CMA-ES with margin updates the mean vector using $\vb_\mathrm{new}$ (after discretization). 

Then, the candidate solution $\bar{\vb}_\mathrm{new} = \enc(\vb_\mathrm{new})$ is evaluated on the objective function $f$ to be minimized, and the smoothed success rate $\psucc^{(t)} \in [0,1]$ is updated as
\begin{align}
\psucc^{(t+1)} = (1 - c_p) \psucc^{(t)} + c_p \mathbb{I}\{ f( \bar{\vb}_\mathrm{new} ) \leq f( \mv^{(t)} ) \} \enspace,
\label{eq:ecmaes-successrate-update}
\end{align}
where $c_p > 0$ is the smoothing factor. The initial value of the smoothed success rate $\psucc^{(0)}$ is given by the target success rate $\ptarget \in [0,1]$. Based on the $1/5$-success rule~\cite{rechenberg:1973:book}, the (1+1)-CMA-ES with margin updates the step-size as
\begin{align}
\sigma^{(t+1)} = \sigma^{(t)} \exp\left( \frac{1}{d_\sigma} \cdot \frac{\psucc^{(t+1)} - \ptarget}{1 - \ptarget} \right) \enspace.
\label{eq:ecmaes-stepsize-update}
\end{align}

The (1+1)-CMA-ES with margin updates the evolution path $\p_c^{(t)}$, covariance matrix $\cov^{(t)}$, and mean vector $\mv^{(t)}$ when the evaluation value $f( \bar{\vb}_\mathrm{new} )$ of new candidate solution is not inferior to the best evaluation value $f( \mv^{(t)} )$ so far. 
The update rules of $\p_c^{(t)}$ and $\cov^{(t)}$ are given by
\begin{align}
\p_c^{(t+1)} &= (1 - c_c) \p_c^{(t)} + h^{(t+1)} \sqrt{c_c (2 - c_c)} \, \y_\mathrm{new} 
\label{eq:ecmaes-epath-update} \\
\cov^{(t+1)} &= \left( 1 - c_1 + \delta(h^{(t+1)}) \right) \cov^{(t)} + c_1 \p_c^{(t+1)} \left( \p_c^{(t+1)} \right)^\T \enspace,
\label{eq:ecmaes-cov-update}
\end{align}
where $h^{(t+1)} = \mathbb{I}{\{ \psucc^{(t+1)} < p_\mathrm{thresh} \}}$ and $\delta(h) = (1 - h) c_1 c_c (2 - c_c)$.
The value $h^{(t+1)}$ stalls the update of the evolution path when the smoothed success rate is larger than a threshold $p_\mathrm{thresh}$. This prevents a fast increase of axes of the covariance matrix when the step-size is too small. 
Note that, introducing the update based on Cholesky decomposition of the covariance matrix~\cite{Suttorp:2009}, the computational cost required in a single update, including the margin correction explained later, can be reduced to $O(N^2)$.

As the candidate solution used in the update of the mean vector, we use the discretized elitist solution $\bar{\vb}_\mathrm{new}$, i.e., we update the mean vector as $\mv^{(t+1)} = \bar{\vb}_\mathrm{new}$ when $\bar{\vb}_\mathrm{new}$ has the best evaluation value. In Section~\ref{sec:ecmaeswm-elitist}, we show that this is a reasonable update rule for the mean vector on the mixed-integer optimization problems.

After the update of distribution parameters, the (1+1)-CMA-ES with margin updates $\A^{(t)}$ by the margin correction. We slightly modify the margin correction to use the encoded elitist solution as the mean vector as explained in Section~\ref{sec:ecmaeswm-margin}.

\begin{algorithm}[t] 
\caption{The (1+1)-CMA-ES with margin}
\begin{algorithmic}[1] \label{alg:ecmaeswm}
\REQUIRE The objective function $f$ to be minimized
\REQUIRE $\mv^{(0)}, \cov^{(0)}, \sigma^{(0)}, \A^{(0)}$ 
\WHILE{termination conditions are not met}
\STATE Generate $\y_\mathrm{new} = ( \cov^{(t)} )^{\frac{1}{2}} \z$ with $\z \sim \N(\mathbf{0}, \I)$. 
\STATE Compute $\vb_\mathrm{new} = \mv^{(t)} + \sigma^{(t)} \A^{(t)} \y_\mathrm{new}$.
\STATE Discretize $\vb_\mathrm{new}$ as $\bar{\vb}_\mathrm{new} = \enc(\vb_\mathrm{new})$.
\STATE Evaluate $f( \bar{\vb}_\mathrm{new} )$.
\STATE Update the smoothed success rate $\psucc^{(t)}$ by \eqref{eq:ecmaes-successrate-update}.
\STATE Update the step-size $\sigma^{(t)}$ by \eqref{eq:ecmaes-stepsize-update}.
\IF{$f( \bar{\vb}_\mathrm{new} ) \leq f( \mv^{(t)} )$}
\STATE Update the mean vector as $\mv^{(t+1)} = \bar{\vb}_\mathrm{new}$. \label{step:ecmaes-update-mean}
\STATE Update $\p_c^{(t)}$ and $\cov^{(t)}$ by \eqref{eq:ecmaes-epath-update} and \eqref{eq:ecmaes-cov-update}, respectively.
\ELSE
\STATE Maintain $\mv^{(t+1)}, \p_c^{(t+1)}, \cov^{(t+1)}$ as $\mv^{(t)}, \p_c^{(t)}, \cov^{(t)}$, respectively.
\ENDIF
\STATE Update $\A^{(t)}$ by margin correction explained \\ in Section~\ref{sec:ecmaeswm-margin}.
\IF{$f$ is on the binary or integer domain}
\STATE Modify $\sigma^{(t+1)}$ and $\A^{(t+1)}$ by the post-process \\ explained in Section~\ref{sec:ecmaeswm-postprocess}.
\ENDIF
\STATE $t \leftarrow t+1$
\ENDWHILE
\end{algorithmic} 
\end{algorithm}
\begin{figure*}[!t]
\centering
\includegraphics[width=0.95\linewidth]{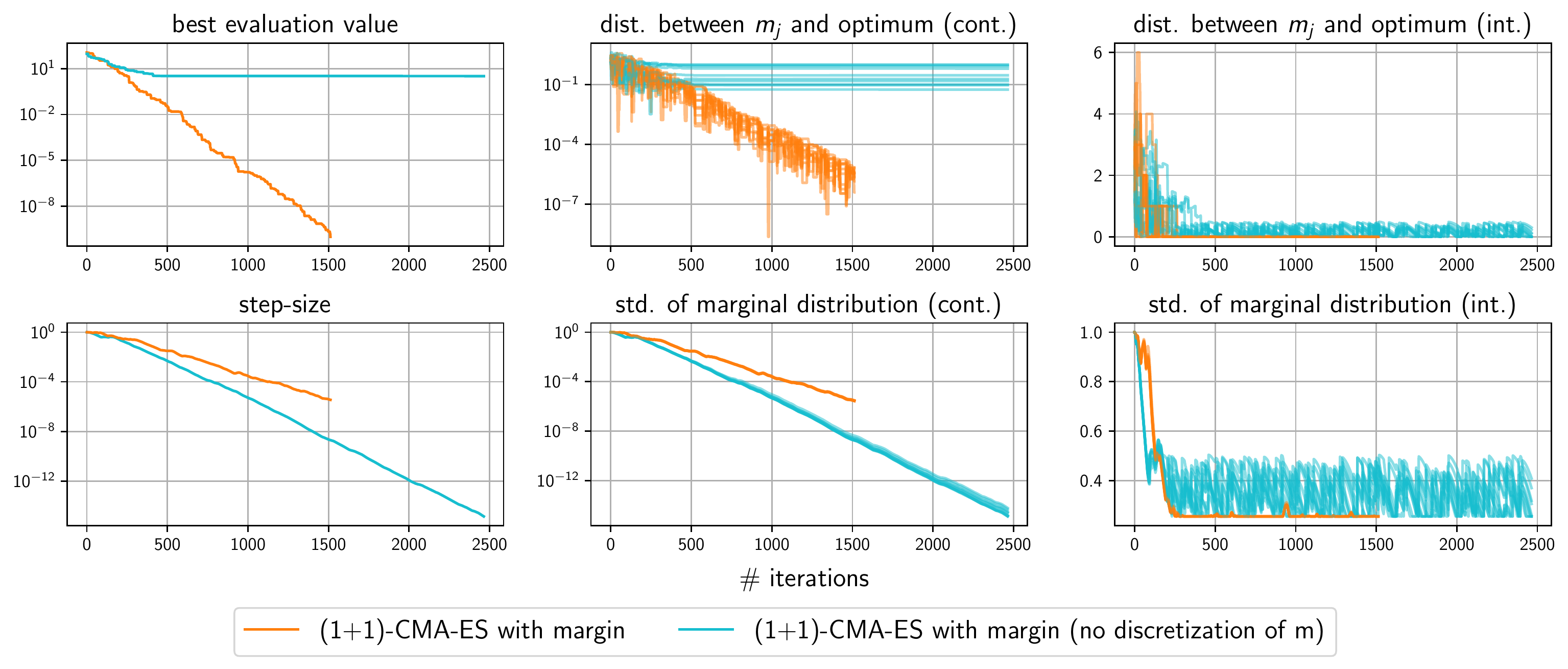}
\caption{Comparison of the (1+1)-CMA-ES with margin with and without discretization of the mean vector.}
\label{fig:ecmaes-comp}
\end{figure*}
\begin{table}[t]
\centering
\caption{Recommended hyperparameter setting of the (1+1)-CMA-ES with margin.}
\begin{tabular}{l}
\hline
    step-size adaptation: \\
    \hspace{10pt} $d_\sigma = 1 + \dfrac{N}{2}$ , \hspace{10pt}
    $\ptarget = \dfrac{2}{11}$ , \hspace{10pt} 
    $c_p = \dfrac{1}{12}$ \hspace{10pt}\\
    covariance matrix adaptation: \\
    \hspace{10pt} $c_c = \dfrac{2}{N+2}$ , \hspace{10pt}
    $c_1 = \dfrac{2}{N^2+6}$ , \hspace{10pt} 
    $p_\mathrm{thresh} = 0.44$ \hspace{10pt}\\
    margin parameter: \\
    \hspace{10pt} $\alpha = \dfrac{1}{N}$ \\
\hline
\end{tabular}
\label{table:hyperparameter-ecmaeswm}
\end{table}
%
%
%

\subsection{Discretization of Mean Vector} \label{sec:ecmaeswm-elitist}

As another possible choice for the updated value of the mean vector $\mv^{(t+1)}$ in success, one can consider the candidate solution $\vb_\mathrm{new}$ before discretization rather than $\bar{\vb}_\mathrm{new}$. To compare the search performance, we optimized the $20$-dimensional \textsc{SphereInt} function (defined in Section~\ref{sec:experiment-mixed}) with $N_\mathrm{co} = N_\mathrm{in} = 10$ by two $(1+1)$-CMA-ESs with margin. One is explained in Section~\ref{sec:ecmaeswm-overall}, and the other updates the mean vector as $\mv^{(t+1)} = {\vb}_\mathrm{new}$ in success. 

Figure~\ref{fig:ecmaes-comp} shows the transitions of the best evaluation value $f(\mv^{(t)})$, step-size $\sigma^{(t)}$, coordinate distances $|\ele{\mv^{(t)}}_j|$ between the mean vector and optimal solution $\x_\mathrm{opt} = \mathbf{0}$, and standard deviations $\sigma^{(t)} \dele{ \A^{(t)} }_j \sqrt{ \dele{ \cov^{(t)} }_j }$ of marginal distributions. They were observed in a single typical trial\del{on 20-dimensional SphereInt function (defined in Section~\ref{sec:experiment-mixed})}{} for each method, where a trial was terminated when the best evaluation value reached $10^{-10}$ or the smallest eigenvalue of $(\sigma^{(t)})^2 \cov^{(t)}$ became smaller than $10^{-30}$. 
In the case without discretization, we observe that the best evaluation value and the elements of the mean vector corresponding to the continuous variables were stalled due to the decrease of the standard deviations of marginal distributions. 
When focusing on the dimensions corresponding to the integer variables, the elements of the mean vector often move close to the midpoints, and the standard deviations of marginal distributions are large. Due to this behavior, the generation probability of other integers becomes too high after the integer variables reach their optimal solutions, and the step-size decreases rapidly by the success-based step-size adaptation. 
In contrast, with the discretization of the mean vector, the mean vector stayed away from the closest midpoints, and the standard deviations of marginal distribution were relatively low.
The discretization of the mean vector maintains the generation probability of other integers around the margin parameter $\alpha$ and realizes the effective search performance on mixed-integer optimization problems.

\begin{figure*}[t]
  \centering
  \includegraphics[width=0.75\linewidth]{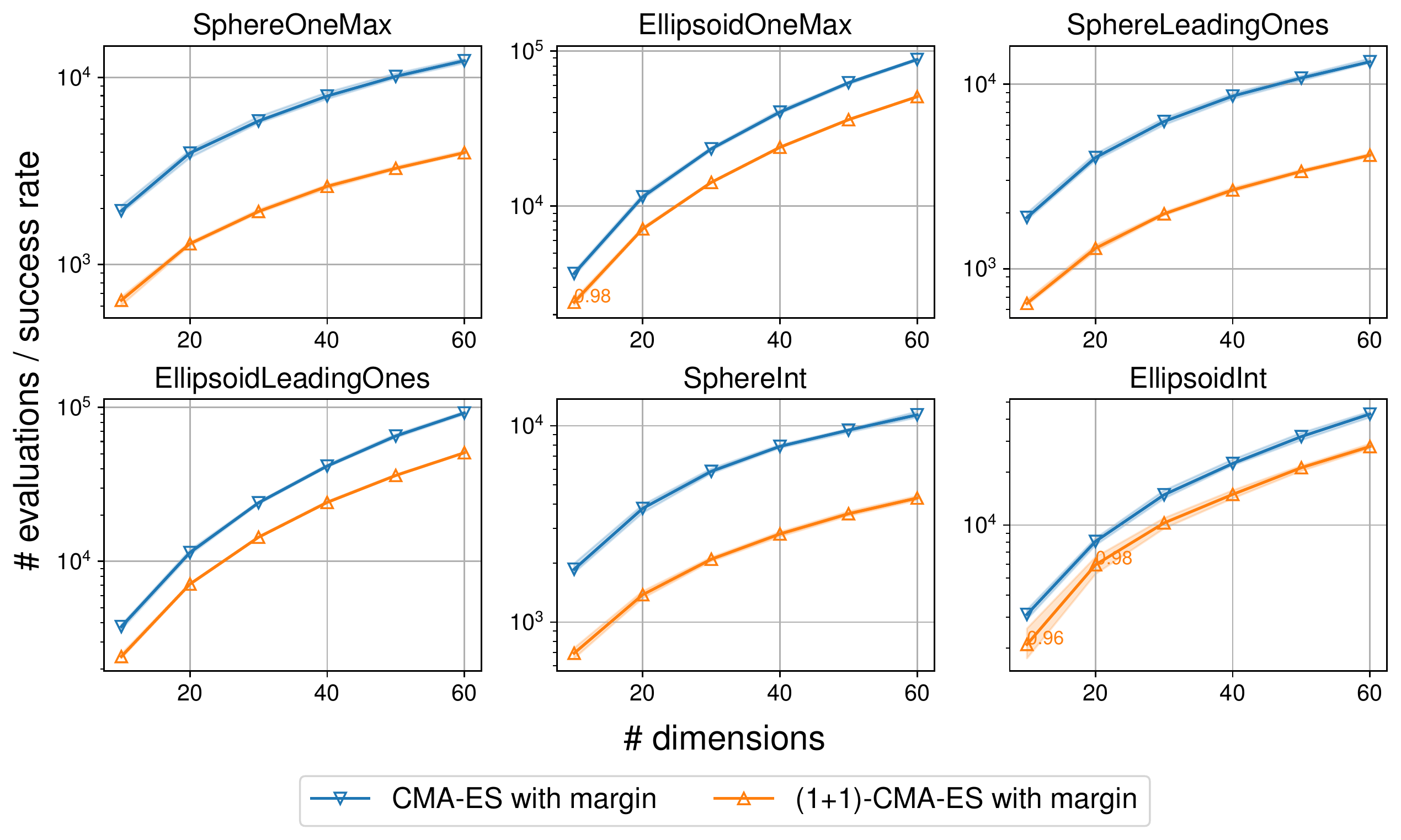}
  \caption{The medians and interquartile ranges of the number of evaluations on the mixed-integer optimization problems. The success rates are also shown if they are not one.}
  \label{fig:result:mixed}
\end{figure*}
%
%
%

\subsection{Margin Correction with Discretized Mean Vector} \label{sec:ecmaeswm-margin}
Since the original margin correction in Section~\ref{sec:cmaeswm-correction} moves the updated mean vector, it reduces the effect of the usage of the encoded mean vector. Therefore, we refine the margin correction not to move the mean vector. Similar to the original margin correction, the modified margin correction applies different update rules of $\A^{(t)}$, which is determined by whether the element of the mean vector is the smallest integer or largest integer, or others.

\paragraph{Case of Smallest or Largest Integer}
Instead of the mean vector, we update the matrix $\A^{(t)}$ to satisfy $\Pr( \ele{\bar{\vb}}_j \neq \ele{{\mv}^{(t+1)}}_j ) \geq \alpha$. This is satisfied when the distance between the mean vector $[\mv^{(t+1)}]_j$ and the nearest midpoint $\ell( \ele{\mv^{(t+1)}}_j )$ in $j$-th dimension is smaller than the confidence interval, i.e.,
\begin{align}
\left| \ele{{\mv}^{(t+1)}}_j - \ell( \ele{\mv^{(t+1)}}_j ) \right| \leq \ci^{(t+1)}_{j, 1-2\alpha} \enspace. \label{eq:ecmaes-margin:edge-cond}
\end{align}
To satisfy \eqref{eq:ecmaes-margin:edge-cond}, when it is not satisfied, $\A^{(t)}$ is updated as
\begin{align}
\dele{\A^{(t+1)}}_j = \frac{ \left| \ele{{\mv}^{(t+1)}}_j - \ell( \ele{\mv^{(t+1)}}_j ) \right| }{ \sigma^{(t+1)} \sqrt{ \dele{ \cov^{(t+1)} }_j  \chi^2_\mathrm{ppf}(1 - 2\alpha) }} \enspace.
\end{align}
We update \del{$\dele{\A^{(t+1)}}_j$ to $\dele{\A^{(t)}}_j$}{}\new{$\dele{\A^{(t)}}_j$ to $\dele{\A^{(t+1)}}_j$} when \eqref{eq:ecmaes-margin:edge-cond} is satisfied before the margin correction.

\paragraph{Case of Other Integers}
We assume the integers $z_{j,1}, \cdots z_{j,K_j}$ for integer variable in $j$-th dimension are at even intervals, i.e., it satisfies $z_{j,k} - z_{j,k+1} = z_{j,k'} - z_{j,k'+1}$ for all $k, k' \in \{1, \cdots, K_j-1\}$.\footnote{\new{This assumption can be easily satisfied by introducing a order-preserving bijective mapping from $\mathcal{Z}_j$ to evenly-spaced integers. We note that, however, such mapping may change the problem characteristics.}}
Then, with the discretized mean vector, $\plow$ and $\pup$ in \eqref{eq:cmeswm-margin:plow} and \eqref{eq:cmeswm-margin:pup} take the same value. As a result, the mean vector is not moved by the correction in~\eqref{eq:cmeswm-margin:mean}. Therefore, we update the matrix $\A^{(t)}$ by the same update rule with the CMA-ES with margin as \eqref{eq:cmeswm-margin:A}.

\subsection{Hyperparameter Setting} \label{sec:ecmaeswm-hyperparameter}
Table~\ref{table:hyperparameter-ecmaeswm} shows the recommended hyperparameter setting of (1+1)-CMA-ES with margin.
To maintain the search performance on the continuous optimization problems, we inherit the default hyperparameter setting of (1+1)-CMA-ES~\cite{Igel:2006} except for the margin parameter. For the margin parameter, we use the default setting of $\alpha = 1/N$ of the CMA-ES with margin with $\lambda = 1$. 

\subsection{Post-process for Discrete Optimization} \label{sec:ecmaeswm-postprocess}
In binary and integer optimizations, the covariance matrix $(\sigma^{(t)})^2 \cov^{(t)}$ often converges rapidly. The (1+1)-CMA-ES with margin (and the CMA-ES with margin) can deal with such convergence by the increase of $\A^{(t)}$. However, due to the numerical error, the behavior of the (1+1)-CMA-ES with margin becomes unstable. To address this problem, we introduce a post-process to reduce the numerical error without changing the \del{behavior}{}\new{algorithm} of the (1+1)-CMA-ES with margin in principle.
We change \del{the diagonal element of $\A^{(t+1)}$ and step-size}{}\new{the step-size and diagonal elements of $\A^{(t+1)}$} after the margin correction as
\begin{align}
    \sigma^{(t+1)} &\leftarrow \sigma^{(t+1)} \cdot \min_{k = 1, \cdots, N} \dele{ \A^{(t+1)} }_k \enspace.
    \label{eq:post-integer-1} \\
    \dele{ \A^{(t+1)} }_j &\leftarrow \frac{ \dele{ \A^{(t+1)} }_j }{ \min_{k = 1, \cdots, N} \dele{ \A^{(t+1)} }_k } \quad \text{for } j=1, \cdots, N 
    \label{eq:post-integer-2} 
\end{align}
This post-process maintains the covariance of $\vb$ in the next iteration and does not change \del{the behavior of the (1+1)-CMA-ES with margin}{}\new{the algorithm but only changes the implementation}.

\begin{figure*}[t]
  \centering
  \includegraphics[width=0.86\linewidth]{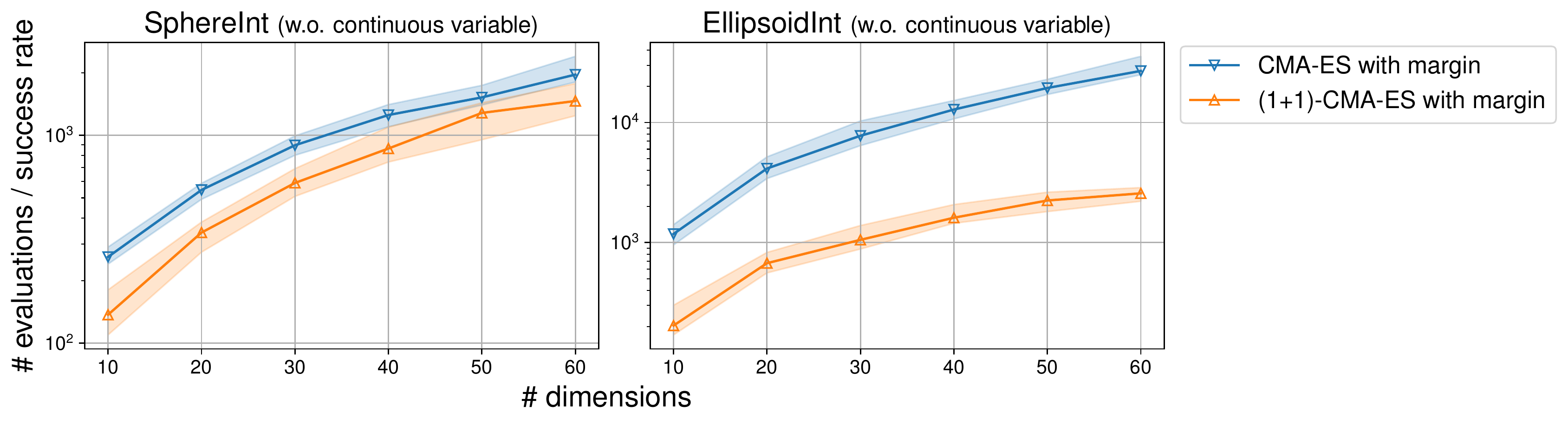}
  \caption{The medians and interquartile ranges of the number of evaluations over 50 independent trials on the integer optimization problems. We note that all trials were successful.}
  \label{fig:result:int}
\end{figure*}
\begin{figure*}[t]
  \centering
  \includegraphics[width=\linewidth]{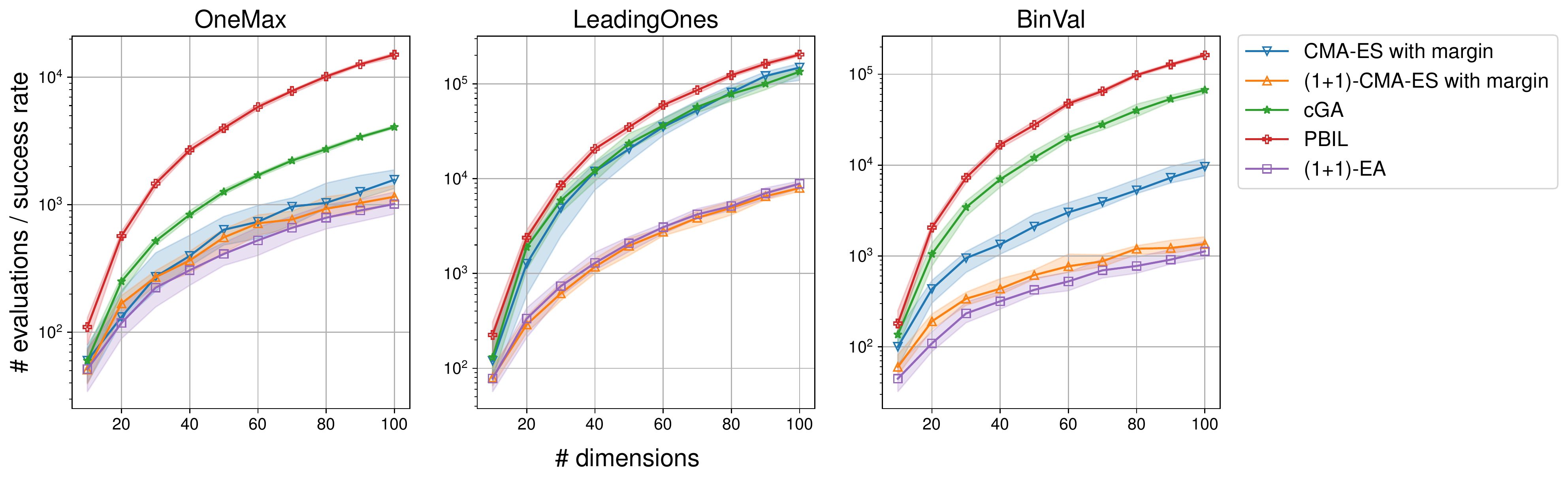}
  \caption{The medians and interquartile ranges of the number of evaluations over 50 independent trials on the binary optimization problems. We note that all trials were successful.}
  \label{fig:result:bi}
\end{figure*}
%
%
%

\section{Experiment} \label{sec:experiment}
We evaluated the search performance of the CMA-ES with margin and the $(1+1)$-CMA-ES with margin on the mixed-integer, integer, and binary domains. We note that the hyperparameters of these algorithms are set as their default values.

\subsection{Experiment on Mixed-Integer Optimization} \label{sec:experiment-mixed}

\paragraph{Experimental Setting}
We used the following benchmark functions to be minimized, that were used in~\cite{cmaeswm}.
\begin{itemize}
\item $\textsc{SphereOneMax}(\bar{\vb}) = \sum^{N_\mathrm{co}}_{j=1} \ele{ \bar{\vb} }_j^2 + N_\mathrm{bi} - \sum^{N}_{j=N_\mathrm{co}+1} \ele{ \bar{\vb} }_j$
\item $\textsc{SphereLeadingOnes}(\bar{\vb}) = $\\
\hspace*{0pt} $\sum^{N_\mathrm{co}}_{j=1} \ele{ \bar{\vb} }_j^2 + N_\mathrm{bi} - \sum^{N}_{j=N_\mathrm{co}+1} \left( \prod^j_{k=N_\mathrm{co}+1} \ele{ \bar{\vb} }_k \right)$
\item $\textsc{EllipsoidOneMax}(\bar{\vb}) = $\\
\hspace*{0pt} $\sum^{N_\mathrm{co}}_{j=1} \Bigl( 1000^{\frac{j-1}{N_\mathrm{co}-1}} \ele{ \bar{\vb} }_j \Bigr)^2 + N_\mathrm{bi} - \sum^{N}_{j=N_\mathrm{co}+1} \ele{ \bar{\vb} }_j$
\item $\textsc{EllipsoidLeadingOnes}(\bar{\vb}) = $\\
\hspace*{0pt} $\sum^{N_\mathrm{co}}_{j=1} \Bigl( 1000^{\frac{j-1}{N_\mathrm{co}-1}} \ele{ \bar{\vb} }_j \Bigr)^2 + N_\mathrm{bi} - \sum^{N}_{j=N_\mathrm{co}+1} \left( \prod^j_{k=N_\mathrm{co}+1} \ele{ \bar{\vb} }_k \right)$
\item $\textsc{SphereInt}(\bar{\vb}) = \sum^{N}_{j=1} \ele{ \bar{\vb} }_j^2$
\item $\textsc{EllipsoidInt}(\bar{\vb}) = \sum^{N}_{j=1} \Bigl( 1000^{\frac{j-1}{N-1}} \ele{ \bar{\vb} }_j \Bigr)^2$
\end{itemize}
The search space of the first four benchmark functions contains $N_\mathrm{co}$ continuous variables and $N_\mathrm{bi}$ binary variables. The last two functions have $N_\mathrm{co}$ continuous variables and $N_\mathrm{in}$ integer variables that can take the integers in $[-10,10]$. We varied the number of dimensions as $N = 10, 20, \cdots, 60$ and set $N_\mathrm{co} = N_\mathrm{bi} = N_\mathrm{in} = N/2$. We performed 50 independent trials for each experimental setting. A trial was considered successful when the best evaluation value reached $10^{-10}$ before the number of evaluations reached $N \times 10^5$ or before the minimal eigenvalue of $(\sigma^{(t)})^2 \cov^{(t)}$ became less than $10^{-30}$. The element of the initial mean vector was given by $\ele{\mv^{(0)}}_j = 0.5$ for binary variables and given uniformly at random in $[1,3]$ otherwise. Other distribution parameters were initialized as $\sigma^{(0)} = 1$ and $\cov^{(0)} = \I$.

\paragraph{Experimental Result}
Figure~\ref{fig:result:mixed} shows the medians and interquartile ranges of the number of evaluations over successful trials divided by the success rate. Figure~\ref{fig:result:mixed} also shows the success rate when there was at least one unsuccessful trial. We can see that the (1+1)-CMA-ES with margin outperformed the CMA-ES with margin on all benchmark functions. We consider the elitist strategy works effectively because the benchmark functions are unimodal. However, on \textsc{EllipsoidOneMax} and \textsc{EllipsoidInt}, the (1+1)-CMA-ES with margin sometimes failed due to the premature convergence of the continuous variables, as well as discussed in Section~\ref{sec:ecmaeswm-elitist}. One possible reason for this is that the default setting of the margin parameter $\alpha = 1 / N$ is too large for low-dimensional mixed-integer optimization problems.

\begin{figure*}[t]
\centering
\includegraphics[width=1\linewidth]{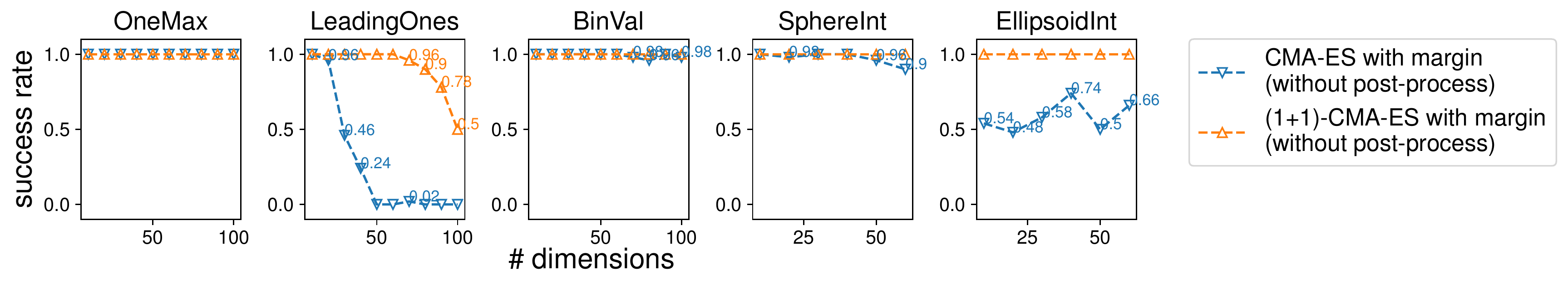}
\caption{The success rate of the CMA-ES with margin and the (1+1)-CMA-ES with margin not applied post-process on the discrete optimization problems. The success rates were computed over 50 independent trials.}
\label{fig:result:post}
\end{figure*}
%
%
%

\subsection{Experiment on Integer Optimization} \label{sec:experiment-int}

\paragraph{Experimental Setting}
We use the \textsc{SphereInt} and \textsc{EllipsoidInt} defined in Section~\ref{sec:experiment-mixed} with $N_\mathrm{co} = 0$ and $N_\mathrm{in} = N$ as the benchmark functions. We varied the number of dimensions as $N = 10, 20, \cdots, 60$ and performed 50 independent trials for each experimental setting. The terminate conditions and initial distribution parameters were set as in Section~\ref{sec:experiment-mixed}. A trial was considered successful if the optimal solution was found before the terminate conditions were met.

We applied the post-process explained in Section~\ref{sec:ecmaeswm-postprocess} to both (1+1)-CMA-ES with margin and CMA-ES with margin. The effect of post-process on the discrete domain will be discussed in Section~\ref{sec:experiment-post}.

\paragraph{Experimental Result}
Figure~\ref{fig:result:int} shows the medians and interquartile ranges of the number of evaluations over successful trials. We note that all the trials were successful. We can see that the (1+1)-CMA-ES with margin outperformed the CMA-ES with margin on both functions. Unlike in the result of mixed-integer optimization problems in Figure~\ref{fig:result:mixed}, the performance gap between the (1+1)-CMA-ES with margin and CMA-ES with margin on \textsc{EllipsoidInt} was larger than that on \textsc{SphereInt}. We consider the reason as follows. Thanks to the elitist strategy and discretization of the mean vector, the (1+1)-CMA-ES with margin can move its mean vector quickly on \textsc{EllipsoidInt}, even before the adaptation of the covariance matrix. In contrast, \del{the CMA-ES with margin adapts the covariance matrix first, then moves the mean vector toward the optimal solution}{}\new{the mean vector of the CMA-ES with margin is not updated efficiently until the covariance matrix is adapted}. This shows the effectiveness of the elitist strategy on the integer domain.

\subsection{Experiment on Binary Optimization} \label{sec:experiment-bin}

\paragraph{Experimental Setting}
We used the following benchmark functions to be maximized.
\begin{itemize}
\item $\textsc{OneMax}(\bar{\vb}) = \sum^{N}_{j=1} \ele{ \bar{\vb} }_j$
\item $\textsc{LeadingOnes}(\bar{\vb}) = \sum^{N}_{j=1} \left( \prod^j_{k=1} \ele{ \bar{\vb} }_k \right)$
\item $\textsc{BinVal}(\bar{\vb}) = \sum^{N}_{j=1} 2^{N-j} \ele{ \bar{\vb} }_j$
\end{itemize}
We varied the number of dimensions as $N = 10, 20, \cdots, 100$ and performed 50 independent trials for each experimental setting. The terminate conditions and initial distribution parameters were set as in Section~\ref{sec:experiment-mixed}. A trial was considered successful if the optimal solution was found before the terminate conditions were met.
We compared the (1+1)-CMA-ES with margin to the binary optimization methods, including the compact genetic algorithm (cGA), population-based incremental learning (PBIL), (1+1)-evolutionary algorithm ((1+1)-EA), in addition to the CMA-ES with margin. We set the sample size of PBIL as the default population size $\lambda = 4 + \lfloor 3 \ln N \rfloor$ of the CMA-ES with margin. \del{We set the learning rates for the cGA and PBIL to $1/N$. The mutation rate of (1+1)-EA was also set to $1/N$.}{}
\new{The mutation rate of (1+1)-EA was set to $1/N$. \nnew{In the cGA and PBIL,} we set the learning rates to $1/N$ and assigned margins of $1/N$ \nnew{for each dimension to leave the possibility of changing each variable}. We note that the setting of learning rate may change the search performance of PBIL, as investigated in~\cite{pbil:lr-effect}.}

As well as the integer case, the covariance $(\sigma^{(t)})^2 \cov^{(t)}$ of the CMA-ES with margin and the (1+1)-CMA-ES with margin often converges rapidly. For the (1+1)-CMA-ES with margin, the post-process in \eqref{eq:post-integer-1} and \eqref{eq:post-integer-2} can prevent such convergence.
In the CMA-ES with margin on the binary domain, since the post-process in \eqref{eq:post-integer-1} and \eqref{eq:post-integer-2} cannot prevent the convergence, we changed the mean vector and step-size after the margin correction as
\begin{align}
    \ele{\mv^{(t+1)}}_j &\leftarrow \frac{ \ele{\mv^{(t+1)}}_j - \ell_{1|2,j} }{ \sigma^{(t+1)} } + \ell_{1|2,j} \quad \text{for } j=1, \cdots, N 
    \label{eq:post-binary-1}\\
    \sigma^{(t+1)} &\leftarrow 1 \enspace.
    \label{eq:post-binary-2}
\end{align}
We note $\ell_{1|2,j}$ is set to $0.5$ on the binary domain.
This post-process also preserves the probability distribution of $\bar{\vb}$ and does not change the \del{behavior}{}\new{algorithm} of the CMA-ES with margin in principle.

\paragraph{Experimental Result}
Figure~\ref{fig:result:bi} shows the medians and interquartile ranges of the number of evaluations over successful trials. We note that all the trials were successful. When focusing the (1+1)-CMA-ES with margin, it is competitive to the (1+1)-EA and achieves the first- or second-best performance on all functions. \del{This result implies the (1+1)-CMA-ES with margin can effectively optimize the mixed-integer optimization problem with $N_\mathrm{co} \ll N_\mathrm{bi}$, which (1+1)-EA cannot be applied to directly.}{}\new{This result implies that the (1+1)-CMA-ES with margin is a reasonable choice for mixed-integer optimization problem with $N_\mathrm{co} \ll N_\mathrm{bi}$, which (1+1)-EA cannot be applied to directly.}
Comparing the PBIL and CMA-ES with margin shows that the CMA-ES with margin outperformed the PBIL on all functions. When the evaluations of candidate solutions can be performed in parallel, the CMA-ES with margin is beneficial in the binary domain.

\subsection{Experiment without Post-Process} \label{sec:experiment-post}
As an abbreviation, we performed the CMA-ES with margin and the (1+1)-CMA-ES with margin without \del{port-process}{}\new{their post-processes} on binary and integer benchmark functions. The terminate conditions and initial distribution parameters were set as in previous sections. Figure~\ref{fig:result:post} shows the success rates computed over 50 independent trials. We note that all trials were successful when applying the post-process, as shown in Section~\ref{sec:experiment-int} and Section~\ref{sec:experiment-bin}. We can confirm that some trails failed by the convergence of $(\sigma^{(t)})^2 \cov^{(t)}$ on some functions, especially on \textsc{LeadingOnes} and \textsc{EllipsoidInt}.
We also confirmed that the medians and interquartile ranges of the numbers of evaluations in successful trials were almost the same as the cases with post-process. This reveals the necessity of the post-process on discrete optimization problems.

\section{Conclusion}
We proposed the (1+1)-CMA-ES with margin, which is derived by introducing the margin correction into (1+1)-CMA-ES. To prevent the premature convergence of continuous variables in mixed-integer optimization problems, we introduced the discretization of the mean vector and modified the margin correction not to move the mean vector. We also applied the post-process for binary and integer optimizations so that the behavior is not affected by the numerical errors. The experimental results on mixed-integer, integer, and binary domains show that the (1+1)-CMA-ES with margin outperforms the CMA-ES with margin.
In the result on the binary domain, in particular, the (1+1)-CMA-ES with margin achieves the first- or second-best performance among the well-known binary optimization methods.

There are a lot of additional components which specialize the CMA-ES to particular situations, such as constraint handling.
The development of novel discrete and mixed-integer optimization methods by transferring the components for CMA-ES to the (1+1)-CMA-ES with margin is one of our future works.
In addition, considering a few unsuccessful trials on some low-dimensional mixed-integer benchmark functions, the investigation of the relationship between the optimization performance and the hyperparameter setting is necessary to provide a more reliable \del{recommended}{}default hyperparameter setting\del{ depending on the number of continuous, integer, binary variables $N_\mathrm{co}, N_\mathrm{in}, N_\mathrm{bi}$}{}.

\begin{acks}
This work was partially supported by JSPS KAKENHI (JP20J23664, JP20H04240), NEDO (JPNP18002, JPNP20006), and JST PRESTO (JPMJPR2133).
\end{acks}

\bibliographystyle{ACM-Reference-Format}
\bibliography{sample-base}


\begin{thebibliography}{21}


\ifx \showCODEN    \undefined \def \showCODEN     #1{\unskip}     \fi
\ifx \showDOI      \undefined \def \showDOI       #1{#1}\fi
\ifx \showISBNx    \undefined \def \showISBNx     #1{\unskip}     \fi
\ifx \showISBNxiii \undefined \def \showISBNxiii  #1{\unskip}     \fi
\ifx \showISSN     \undefined \def \showISSN      #1{\unskip}     \fi
\ifx \showLCCN     \undefined \def \showLCCN      #1{\unskip}     \fi
\ifx \shownote     \undefined \def \shownote      #1{#1}          \fi
\ifx \showarticletitle \undefined \def \showarticletitle #1{#1}   \fi
\ifx \showURL      \undefined \def \showURL       {\relax}        \fi
\providecommand\bibfield[2]{#2}
\providecommand\bibinfo[2]{#2}
\providecommand\natexlab[1]{#1}
\providecommand\showeprint[2][]{arXiv:#2}

\bibitem[\protect\citeauthoryear{Arnold and Hansen}{Arnold and Hansen}{2012}]%
        {cmaes:constraint3}
\bibfield{author}{\bibinfo{person}{Dirk~V. Arnold} {and}
  \bibinfo{person}{Nikolaus Hansen}.} \bibinfo{year}{2012}\natexlab{}.
\newblock \showarticletitle{A {(1+1)-CMA-ES} for Constrained Optimisation}. In
  \bibinfo{booktitle}{\emph{Proceedings of the 14th Annual Conference on
  Genetic and Evolutionary Computation}}. \bibinfo{publisher}{Association for
  Computing Machinery}, \bibinfo{address}{New York, NY, USA},
  \bibinfo{pages}{297--304}.
\newblock
\showISBNx{9781450311779}
\urldef\tempurl%
\url{https://doi.org/10.1145/2330163.2330207}
\showDOI{\tempurl}


\bibitem[\protect\citeauthoryear{Baluja}{Baluja}{1994}]%
        {pbil}
\bibfield{author}{\bibinfo{person}{Shummet Baluja}.}
  \bibinfo{year}{1994}\natexlab{}.
\newblock \bibinfo{booktitle}{\emph{Population-Based Incremental Learning: A
  Method for Integrating Genetic Search Based Function Optimization and
  Competitive Learning}}.
\newblock \bibinfo{type}{{T}echnical {R}eport}. \bibinfo{institution}{Carnegie
  Mellon University Pittsburgh}.
\newblock


\bibitem[\protect\citeauthoryear{Folly and Venayagamoorthy}{Folly and
  Venayagamoorthy}{2009}]%
        {pbil:lr-effect}
\bibfield{author}{\bibinfo{person}{Komla~A. Folly} {and}
  \bibinfo{person}{Ganesh~K. Venayagamoorthy}.}
  \bibinfo{year}{2009}\natexlab{}.
\newblock \showarticletitle{Effects of learning rate on the performance of the
  population based incremental learning algorithm}. In
  \bibinfo{booktitle}{\emph{2009 International Joint Conference on Neural
  Networks}}. \bibinfo{pages}{861--868}.
\newblock
\urldef\tempurl%
\url{https://doi.org/10.1109/IJCNN.2009.5179080}
\showDOI{\tempurl}


\bibitem[\protect\citeauthoryear{Fujii, Takahashi, and Akimoto}{Fujii
  et~al\mbox{.}}{2018}]%
        {mixed-integer-opt:example2}
\bibfield{author}{\bibinfo{person}{Garuda Fujii}, \bibinfo{person}{Masayuki
  Takahashi}, {and} \bibinfo{person}{Youhei Akimoto}.}
  \bibinfo{year}{2018}\natexlab{}.
\newblock \showarticletitle{{CMA}-{ES}-based structural topology optimization
  using a level set boundary expression--{Application} to optical and carpet
  cloaks}.
\newblock \bibinfo{journal}{\emph{Computer Methods in Applied Mechanics and
  Engineering}}  \bibinfo{volume}{332} (\bibinfo{year}{2018}),
  \bibinfo{pages}{624--643}.
\newblock
\showISSN{0045-7825}
\urldef\tempurl%
\url{https://doi.org/10.1016/j.cma.2018.01.008}
\showDOI{\tempurl}


\bibitem[\protect\citeauthoryear{Hamano, Saito, Nomura, and Shirakawa}{Hamano
  et~al\mbox{.}}{2022}]%
        {cmaeswm}
\bibfield{author}{\bibinfo{person}{Ryoki Hamano}, \bibinfo{person}{Shota
  Saito}, \bibinfo{person}{Masahiro Nomura}, {and} \bibinfo{person}{Shinichi
  Shirakawa}.} \bibinfo{year}{2022}\natexlab{}.
\newblock \showarticletitle{CMA-ES with Margin: {L}ower-Bounding Marginal
  Probability for Mixed-Integer Black-Box Optimization}. In
  \bibinfo{booktitle}{\emph{Proceedings of the Genetic and Evolutionary
  Computation Conference}} \emph{(\bibinfo{series}{GECCO '22})}.
  \bibinfo{publisher}{Association for Computing Machinery},
  \bibinfo{address}{New York, NY, USA}, \bibinfo{pages}{639--647}.
\newblock
\showISBNx{9781450392372}
\urldef\tempurl%
\url{https://doi.org/10.1145/3512290.3528827}
\showDOI{\tempurl}


\bibitem[\protect\citeauthoryear{Hansen}{Hansen}{2016}]%
        {hansen:2011:tutorial}
\bibfield{author}{\bibinfo{person}{Nikolaus Hansen}.}
  \bibinfo{year}{2016}\natexlab{}.
\newblock \showarticletitle{The {CMA} Evolution Strategy: {A} Tutorial}.
\newblock \bibinfo{journal}{\emph{arXiv:1604.00772}} (\bibinfo{year}{2016}).
\newblock
\showeprint[arxiv]{1604.00772}


\bibitem[\protect\citeauthoryear{Hansen, M{\"u}ller, and Koumoutsakos}{Hansen
  et~al\mbox{.}}{2003}]%
        {hansen:2003:ec}
\bibfield{author}{\bibinfo{person}{Nikolaus Hansen},
  \bibinfo{person}{Sibylle~D. M{\"u}ller}, {and} \bibinfo{person}{Petros
  Koumoutsakos}.} \bibinfo{year}{2003}\natexlab{}.
\newblock \showarticletitle{Reducing the Time Complexity of the Derandomized
  Evolution Strategy with Covariance Matrix Adaptation ({CMA-ES})}.
\newblock \bibinfo{journal}{\emph{IEEE Transactions on Evolutionary
  Computation}}  \bibinfo{volume}{11} (\bibinfo{year}{2003}),
  \bibinfo{pages}{1--18}.
\newblock


\bibitem[\protect\citeauthoryear{Hansen, Niederberger, Guzzella, and
  Koumoutsakos}{Hansen et~al\mbox{.}}{2009}]%
        {cmaes:noise1}
\bibfield{author}{\bibinfo{person}{Nikolaus Hansen}, \bibinfo{person}{Andr\'{E}
  S.~P. Niederberger}, \bibinfo{person}{Lino Guzzella}, {and}
  \bibinfo{person}{Petros Koumoutsakos}.} \bibinfo{year}{2009}\natexlab{}.
\newblock \showarticletitle{A Method for Handling Uncertainty in Evolutionary
  Optimization With an Application to Feedback Control of Combustion}.
\newblock \bibinfo{journal}{\emph{IEEE Transactions on Evolutionary
  Computation}} \bibinfo{volume}{13}, \bibinfo{number}{1}
  (\bibinfo{year}{2009}), \bibinfo{pages}{180--197}.
\newblock
\urldef\tempurl%
\url{https://doi.org/10.1109/TEVC.2008.924423}
\showDOI{\tempurl}


\bibitem[\protect\citeauthoryear{Hansen and Ostermeier}{Hansen and
  Ostermeier}{1996}]%
        {hansen:1996:cmaes}
\bibfield{author}{\bibinfo{person}{Nikolaus Hansen} {and}
  \bibinfo{person}{Andreas Ostermeier}.} \bibinfo{year}{1996}\natexlab{}.
\newblock \showarticletitle{Adapting arbitrary normal mutation distributions in
  evolution strategies: the covariance matrix adaptation}. In
  \bibinfo{booktitle}{\emph{Proceedings of IEEE International Conference on
  Evolutionary Computation}}. \bibinfo{publisher}{IEEE},
  \bibinfo{pages}{312--317}.
\newblock
\urldef\tempurl%
\url{https://doi.org/10.1109/ICEC.1996.542381}
\showDOI{\tempurl}


\bibitem[\protect\citeauthoryear{Harik, Lobo, and Goldberg}{Harik
  et~al\mbox{.}}{1999}]%
        {cga}
\bibfield{author}{\bibinfo{person}{G.~R. Harik}, \bibinfo{person}{F.~G. Lobo},
  {and} \bibinfo{person}{D.~E. Goldberg}.} \bibinfo{year}{1999}\natexlab{}.
\newblock \showarticletitle{The Compact Genetic Algorithm}.
\newblock \bibinfo{journal}{\emph{IEEE Transactions on Evolutionary
  Computation}}  \bibinfo{volume}{3} (\bibinfo{year}{1999}),
  \bibinfo{pages}{287--297}.
\newblock
Issue 4.
\urldef\tempurl%
\url{https://doi.org/10.1109/4235.797971}
\showDOI{\tempurl}


\bibitem[\protect\citeauthoryear{Hazan, Klivans, and Yuan}{Hazan
  et~al\mbox{.}}{2018}]%
        {mixed-integer-opt:example3}
\bibfield{author}{\bibinfo{person}{Elad Hazan}, \bibinfo{person}{Adam Klivans},
  {and} \bibinfo{person}{Yang Yuan}.} \bibinfo{year}{2018}\natexlab{}.
\newblock \showarticletitle{Hyperparameter optimization: a spectral approach}.
  In \bibinfo{booktitle}{\emph{International Conference on Learning
  Representations (ICLR)}}.
\newblock


\bibitem[\protect\citeauthoryear{Hellwig and Beyer}{Hellwig and Beyer}{2020}]%
        {cmaes:noise2}
\bibfield{author}{\bibinfo{person}{Michael Hellwig} {and}
  \bibinfo{person}{Hans-Georg Beyer}.} \bibinfo{year}{2020}\natexlab{}.
\newblock \showarticletitle{On the steady state analysis of covariance matrix
  self-adaptation evolution strategies on the noisy ellipsoid model}.
\newblock \bibinfo{journal}{\emph{Theoretical Computer Science}}
  \bibinfo{volume}{832} (\bibinfo{year}{2020}), \bibinfo{pages}{98--122}.
\newblock
\showISSN{0304-3975}
\urldef\tempurl%
\url{https://doi.org/10.1016/j.tcs.2018.05.016}
\showDOI{\tempurl}


\bibitem[\protect\citeauthoryear{Igel, Hansen, and Roth}{Igel
  et~al\mbox{.}}{2007}]%
        {cmaes:multiobj1}
\bibfield{author}{\bibinfo{person}{Christian Igel}, \bibinfo{person}{Nikolaus
  Hansen}, {and} \bibinfo{person}{Stefan Roth}.}
  \bibinfo{year}{2007}\natexlab{}.
\newblock \showarticletitle{Covariance Matrix Adaptation for Multi-objective
  Optimization}.
\newblock \bibinfo{journal}{\emph{Evolutionary Computation}}
  \bibinfo{volume}{15}, \bibinfo{number}{1} (\bibinfo{date}{03}
  \bibinfo{year}{2007}), \bibinfo{pages}{1--28}.
\newblock
\showISSN{1063-6560}
\urldef\tempurl%
\url{https://doi.org/10.1162/evco.2007.15.1.1}
\showDOI{\tempurl}


\bibitem[\protect\citeauthoryear{Igel, Suttorp, and Hansen}{Igel
  et~al\mbox{.}}{2006}]%
        {Igel:2006}
\bibfield{author}{\bibinfo{person}{Christian Igel}, \bibinfo{person}{Thorsten
  Suttorp}, {and} \bibinfo{person}{Nikolaus Hansen}.}
  \bibinfo{year}{2006}\natexlab{}.
\newblock \showarticletitle{A Computational Efficient Covariance Matrix Update
  and a {(1+1)-CMA} for Evolution Strategies}. In
  \bibinfo{booktitle}{\emph{Proceedings of the 8th Annual Conference on Genetic
  and Evolutionary Computation}}. \bibinfo{publisher}{Association for Computing
  Machinery}, \bibinfo{address}{New York, NY, USA}, \bibinfo{pages}{453--460}.
\newblock
\showISBNx{1595931864}
\urldef\tempurl%
\url{https://doi.org/10.1145/1143997.1144082}
\showDOI{\tempurl}


\bibitem[\protect\citeauthoryear{Larson, Menickelly, and Wild}{Larson
  et~al\mbox{.}}{2019}]%
        {larson_menickelly_wild_2019}
\bibfield{author}{\bibinfo{person}{Jeffrey Larson}, \bibinfo{person}{Matt
  Menickelly}, {and} \bibinfo{person}{Stefan~M. Wild}.}
  \bibinfo{year}{2019}\natexlab{}.
\newblock \showarticletitle{Derivative-free optimization methods}.
\newblock \bibinfo{journal}{\emph{Acta Numerica}}  \bibinfo{volume}{28}
  (\bibinfo{year}{2019}), \bibinfo{pages}{287--404}.
\newblock
\urldef\tempurl%
\url{https://doi.org/10.1017/S0962492919000060}
\showDOI{\tempurl}


\bibitem[\protect\citeauthoryear{Piermarini and Roma}{Piermarini and
  Roma}{2021}]%
        {integer-opt:example1}
\bibfield{author}{\bibinfo{person}{Christian Piermarini} {and}
  \bibinfo{person}{Massimo Roma}.} \bibinfo{year}{2021}\natexlab{}.
\newblock \showarticletitle{A Simulation-Based Optimization approach for
  analyzing the ambulance diversion phenomenon in an Emergency Department
  network}.
\newblock \bibinfo{journal}{\emph{arXiv:2108.04162}} (\bibinfo{year}{2021}).
\newblock
\urldef\tempurl%
\url{https://doi.org/10.48550/ARXIV.2108.04162}
\showDOI{\tempurl}


\bibitem[\protect\citeauthoryear{Rechenberg}{Rechenberg}{1973}]%
        {rechenberg:1973:book}
\bibfield{author}{\bibinfo{person}{Ingo. Rechenberg}.}
  \bibinfo{year}{1973}\natexlab{}.
\newblock \bibinfo{booktitle}{\emph{Evolutionsstrategie; Optimierung
  technischer Systeme nach Prinzipien der biologischen Evolution. Mit einem
  Nachwort von Manfred Eigen}}.
\newblock \bibinfo{publisher}{Frommann-Holzboog}.
\newblock
\showISBNx{3772803733}


\bibitem[\protect\citeauthoryear{Rios and Sahinidis}{Rios and
  Sahinidis}{2013}]%
        {Rios:2013}
\bibfield{author}{\bibinfo{person}{Luis~Miguel Rios} {and}
  \bibinfo{person}{Nikolaos~V. Sahinidis}.} \bibinfo{year}{2013}\natexlab{}.
\newblock \showarticletitle{Derivative-free optimization: a review of
  algorithms and comparison of software implementations}.
\newblock \bibinfo{journal}{\emph{Journal of Global Optimization}}
  \bibinfo{volume}{56}, \bibinfo{number}{3} (\bibinfo{year}{2013}),
  \bibinfo{pages}{1247--1293}.
\newblock


\bibitem[\protect\citeauthoryear{Sakamoto and Akimoto}{Sakamoto and
  Akimoto}{2022}]%
        {cmaes:constraint2}
\bibfield{author}{\bibinfo{person}{Naoki Sakamoto} {and}
  \bibinfo{person}{Youhei Akimoto}.} \bibinfo{year}{2022}\natexlab{}.
\newblock \showarticletitle{Adaptive Ranking-Based Constraint Handling for
  Explicitly Constrained Black-Box Optimization}.
\newblock \bibinfo{journal}{\emph{Evolutionary Computation}}
  \bibinfo{volume}{30}, \bibinfo{number}{4} (\bibinfo{date}{12}
  \bibinfo{year}{2022}), \bibinfo{pages}{503--529}.
\newblock
\showISSN{1063-6560}
\urldef\tempurl%
\url{https://doi.org/10.1162/evco_a_00310}
\showDOI{\tempurl}


\bibitem[\protect\citeauthoryear{Suttorp, Hansen, and Igel}{Suttorp
  et~al\mbox{.}}{2009}]%
        {Suttorp:2009}
\bibfield{author}{\bibinfo{person}{Thorsten Suttorp}, \bibinfo{person}{Nikolaus
  Hansen}, {and} \bibinfo{person}{Christian Igel}.}
  \bibinfo{year}{2009}\natexlab{}.
\newblock \showarticletitle{Efficient covariance matrix update for variable
  metric evolution strategies}.
\newblock \bibinfo{journal}{\emph{Machine Learning}} \bibinfo{volume}{75},
  \bibinfo{number}{2} (\bibinfo{year}{2009}), \bibinfo{pages}{167--197}.
\newblock
\showISSN{1573-0565}
\urldef\tempurl%
\url{https://doi.org/10.1007/s10994-009-5102-1}
\showDOI{\tempurl}


\bibitem[\protect\citeauthoryear{Zhang, Apley, and Chen}{Zhang
  et~al\mbox{.}}{2020}]%
        {mixed-integer-opt:example1}
\bibfield{author}{\bibinfo{person}{Yichi Zhang}, \bibinfo{person}{Daniel~W.
  Apley}, {and} \bibinfo{person}{Wei Chen}.} \bibinfo{year}{2020}\natexlab{}.
\newblock \showarticletitle{Bayesian {Optimization} for {Materials} {Design}
  with {Mixed} {Quantitative} and {Qualitative} {Variables}}.
\newblock \bibinfo{journal}{\emph{Scientific Reports}} \bibinfo{volume}{10},
  \bibinfo{number}{1} (\bibinfo{year}{2020}), \bibinfo{pages}{4924}.
\newblock
\showISSN{2045-2322}
\urldef\tempurl%
\url{https://doi.org/10.1038/s41598-020-60652-9}
\showDOI{\tempurl}


\end{thebibliography}

\end{document}